\documentclass[letterpaper]{article}
\usepackage{natbib,alifexi,subfigure}

\title{Passive and Driven Trends in the Evolution of Complexity}
\author{Larry Yaeger$^{1}$, Virgil Griffith$^{1,2}$, \and Olaf Sporns$^3$\\
\mbox{}\\
$^1$School of Informatics and $^3$Department of Psychological and Brain Sciences\\
Indiana University, Bloomington, IN 47408\\
$^2$Computation \& Neural Systems, California Institute of Technology, Pasadena, CA 91125\\
larryy@indiana.edu}

\begin{document}
\maketitle

\begin{abstract}
  The nature and source of evolutionary trends in complexity is difficult to assess from the fossil record, and the \emph{driven vs. passive} nature of such trends has been debated for decades.  There are also questions about how effectively artificial life software can evolve increasing levels of complexity.  We extend our previous work demonstrating an evolutionary increase in an information theoretic measure of neural complexity in an artificial life system (Polyworld), and introduce a new technique for distinguishing driven from passive trends in complexity.  Our experiments show that evolution can and does select for complexity increases in a driven fashion, in some circumstances, but under other conditions it can also select for complexity stability.  It is suggested that the evolution of complexity is entirely driven---just not in a single direction---at the scale of species.  This leaves open the question of evolutionary trends at larger scales.
\end{abstract}

\hyphenation{ap-proximated}

\section{Introduction}

The existence of an evolutionary trend towards greater complexity is undeniable, whether one measures complexity by organism size \citep{Cope1871}, distinct cell types \citep{Bonner1988,Valentine1994}, morphology \citep{Thomas1993,McShea1993}, or ecological webs of interaction \citep{Knoll2000}.  Historically, it has often been suggested that such growth is the result of an evolutionary bias towards forms and functions of greater complexity, and a great variety of rationales has been offered for why this should be the case; e.g., \citep{Rensch1960a, Rensch1960b,Waddington1969,Saunders1976,Kimura1983,Katz1987,Bonner1988,Arthur1994,Huynen1996,Newman1998}; see \citet{McShea1991} and \citet{Carroll2001} for reviews.  However, \citet{MaynardSmith1970}, \citet{Raup1973}, \citet{Gould1994}, and others have questioned whether that growth has been the outcome of natural selection or simply, in Maynard Smith's words, the ``obvious and uninteresting explanation" of a sort of random walk away from an immutable barrier of simplicity at the lower extreme---a growth in variance relative to the necessarily low complexity at the origin of life.

\citet{Bedau1997} and \citet{Rechsteiner1999} provide some evidence of an increasing and accelerating ``evolutionary activity" in biological systems not yet demonstrated in artificial life models.  However, other attempts to characterize complexity trends in the fossil record have produced mixed results at best \citep{McShea1996,Heylighen2000,Carroll2001}, leaving us with no clear picture of the influence of natural selection on complexity.  \citet{McShea1994,McShea1996,McShea2001,McShea2005} has, over the years, attempted to clarify (and, where possible, empirically address) the debate, by identifying distinct classes of complexity and, importantly, by distinguishing between ``driven" trends, in which evolution actively selects for complexity, and ``passive" trends, in which increases in complexity are due simply to asymmetric random drift.

Simple computational models of branching species and clade lineages in simple numerical parameter spaces (arbitrary values standing in for complexity, size, or the like) have been used to investigate this distinction between driven and passive evolutionary trends \citep{Raup1973,Raup1974}, however it is not always possible to distinguish a passive system from a weakly driven system \citep{McShea1994}.  Furthermore, the anagenetic component of some of these models, while intended, by definition, to address within-lineage change, is equivalent to branching lineages that effectively compete with one another for parameter space, by requiring branched, descendant lines to replace ancestral lines.  The typical assumption of equal extinction rates across all scales may also unintentionally color the results from these models.  It is common wisdom, for example, that a driven system necessarily implies an increase in the minimum value of whatever parameter is being used to distinguish taxonomic branches \citep{Wagner1996,McShea2001,Carroll2001}, yet an evolutionary system in which fitness at smaller scales is independent of fitness at larger scales could possess a drive towards larger scales without eliminating or even disadvantaging organisms at the lower end of the spectrum.  Indeed, \citet{McShea1994} acknowledges a dramatically lower rate of growth in this minimum in a purely cladogenetic model compared to a mixed anagenetic and cladogenetic model.  One of us [LY] is currently investigating the effects of the uniform extinction rate assumption.

Given the difficulty and ambiguity one encounters when attempting to answer questions about the evolution of complexity from paleontological data or simple branching models, it makes sense to turn to computer models of evolution to address these questions.  \citet{Turney1999,Turney2000} has used a simple evolutionary model to suggest that increasing evolvability is central to progress in evolution and predicts an accelerating increase in biological systems that might correlate with complexity growth.  \citet{Adami2000,Adami2002} has defined complexity as the information that an organism's genome encodes about its environment and used Avida to show that asexual agents in a fixed, single niche always evolve towards greater complexity of this narrowly defined type.

Modern compute power and artificial life methods allow us to rewind the ``tape of life'', as \citet{Gould1989} put it, and let history unfold again and again under slightly or dramatically different influences.  Here we use a method for ``replaying the tape'' that is substantially beyond the perturbed playback Gould envisaged, demonstrating a method for carrying out parallel simulations in which natural selection either does or does not play a part, yet with all other population and genetical statistics being held constant.  This is similar in spirit to the extreme ``behavioral noise'' null model used by \citet{Bedau1995} and subsequent ``neutral shadow models'' \citep{Bedau1998}.  Being able to effectively turn natural selection on and off in this fashion allows us to tease apart and distinguish evolutionarily driven trends from passive trends in a formal, quantitative fashion.

The trend we investigate is a particular information-theoretic measure of complexity \citep{Tononi1994,Lungarella2005}, \textbf{C}, for the neural dynamics of artificial agents in an evolving computational ecology, Polyworld \citep{Yaeger1994}.  In previous work \citep{Yaeger2006} we demonstrated an increasing trend in \textbf{C} in the agents of Polyworld over evolutionary time scales, and were able to relate these increases to increasing structural elaboration of the agents' neural network architectures and an increase in the learning rates employed at the Hebbian synapses in these networks.  We did not, however, address the evolutionary source of these increases, or whether that source should be construed as driven or passive, in the \citet{McShea1996} sense.  Here we use a novel technique that allows us to make such a distinction, and discover that, at least at the scale of single species and ecological niches, evolution of complexity is always driven, but, interestingly, not always driven in the same direction.

\section{Tools and Techniques}

\subsection{Polyworld}

Polyworld \citep{Yaeger1994} is an evolutionary model of an ecology populated with haploid agents, each with a suite of primitive behaviors (move, turn, eat, mate, attack, light, focus) under continuous control of an Artificial Neural Network (ANN) consisting of summing and squashing neurons and synapses that adapt via Hebbian learning. The architecture of the ANN is encoded in the organism's genome, expressed as a number of neural groups of excitatory and inhibitory neurons, with genetically determined synaptic connection densities, topologies, and learning rates.  Input to the ANN consists of pixels from a rendering of the scene from each agent's point of view, like light falling on a retina.  Though agent morphologies are simple and static, agents interact with the world and each other in fairly complex ways, as they replenish energy by seeking out and consuming food and by attacking, killing, and eating other agents.  They reproduce by simultaneous expression of a mating behavior by two collocated agents.

Agent population is normally bounded above and below, but unlike the simulations discussed in \citep{Yaeger2006}, there is no ``smite" function invoked at maximum population, which the authors felt risked introducing a bias associated with the simulator's \emph{ad hoc} heuristic fitness function.  Nor does a minimum population any longer invoke a steady-state GA, which also would necessarily depend upon that fitness function.  Instead, as the population grows towards the upper bound, the amount of energy depleted by all agent behaviors, including neural activity, is increased in a continuous fashion.  Conversely, as the population drops towards the minimum, energy depletion is decreased, and agent lifespans may be artificially extended.  This does not guarantee a viable population (one that sustains its numbers through reproduction), since unsuccessful, unfit agents may be all that remain by the time a failing population bottoms out, but it does provide very effective population control without denying births to agents capable and ``desirous" of doing so, while simultaneously eliminating any possible effects of the now purely informative heuristic fitness function.

\subsection{Complexity}

For our purposes, complexity, \textbf{C}, is computed using a new C++ implementation based on the methods of \citep{Lungarella2005} for approximating the information-theoretic measure of complexity originally developed by \citep{Tononi1994}.  Though non-trivial to derive and implement, the intuition behind \textbf{C} is straightforward:  Cooperation amongst various elements of a network, called \emph{integration} and measured by a multivariate extension to mutual information, increases network complexity, to a point.  But specialization of network subunits, called \emph{segregation} and measured by the difference between maximum and actual integration at different scales, also increases network complexity.  Maximal complexity is achieved in networks that optimally trade off the opposing tensions between integration and segregation---between cooperation and specialization---and maximize both to the extent possible.  The original measure of complexity is given by: \begin{equation}C_N(X) = \sum_{k=1}^{n} { [\ \langle H(X_{j}^{k}) \rangle - \frac{k}{n}H(X)\ ]}\end{equation} where $H(X)$ is the Shannon entropy of the entire system of $n$ variables, $k$ is the size of a subset of variables, and $j$ indicates that the ensemble average $\langle H(X_{j}^{k}) \rangle$ is to be taken over all $n!/(k!(n-k)!)$ combinations of $k$ variables.  The  simplified approximation we use was introduced in \citep{Tononi1998} and explored computationally in \citep{Sporns2000}: \begin{equation}C(X) = H(X) - \sum_{x_i \in X}^{}{H(x_i|X - x_i) }\end{equation} where $H(X)$ is the entropy of the entire system and the $H(x_i|X - x_i)$ terms are the conditional entropy of each of the variables $x_i$ given the entropy of the rest of the system.

\subsection{Natural Selection \emph{vs.} Random Drift}

While probably impossible to eliminate natural selection from an evolving biological ecology, artificial ecologies are more flexible.  In order to distinguish between evolutionarily driven and passive trends, we designed a mode for running our simulator in which natural selection had effectively been eliminated, yet which could be compared directly with runs in which natural selection operated normally.  This was accomplished by implementing a new ``lockstep" mode of operation in Polyworld.  First a simulation is run in the system's normal, natural-selection mode of operation.  During this natural-selection run, the birth and death of every agent is recorded (along with the usual statistics, brain states, etc.).  Then the simulator is run in the lockstep mode, starting from the same initial conditions as the natural-selection run and using the birth and death data recorded during the natural-selection run.  No ``natural" births or deaths are allowed during a lockstep run.  Instead, every time a birth occurred in the original natural-selection run, a birth is forced to occur in the lockstep run, only instead of being produced by the original parents, the birth is produced by two agents chosen at random from the population.  Similarly, whenever a death occurred in the natural-selection run, a death is forced to occur in the lockstep run, only instead of the original agent dying, a random agent is killed and removed from the population.

By so doing, population statistics are forced to be identical between the paired natural-selection and lockstep runs.  As a result, the genetical statistics---number of crossover and mutation operations---are forced to be comparable in the paired runs.  Note that since crossover and mutation are applied to different genomes and since the number of crossover points and the mutation rate are themselves embedded in these genomes \citep{Yaeger1994}, these genetic operations are only statistically comparable between paired runs, not identical.  Similarly, the ``life experiences" of a given agent---its trajectory through the world and the inputs to its visual system---are only comparable statistically between paired runs.  Since the agents' life experiences do impact the values of neural complexity we compute, this could produce extraneous differences between paired runs, but we do not expect this to have any consistent, measurable influence.  The controlling statistics, such as the entropy and mutual information in the visual inputs, are comparable between paired runs, so we expect computed complexity values to be similarly comparable.  While it would be possible to record number of crossover points, mutation rates, agent trajectories, and even sensory inputs during the natural-selection runs and play them back during the lockstep runs, we do not believe this would alter the relevant statistics or the measured outcomes in a substantive manner, and therefore have not made any such attempts.  One could also argue that since complexity is affected by agent behaviors and their resulting sensory inputs, agents in lockstep runs must be able to control their actions in order to obtain valid measures of their neural complexity.

The end result of these machinations is that gene states are subject to natural selection, based on the evolutionary viability---the fitness---of the agents' behaviors, in the natural-selection runs.  While gene states are subject only to the same degree of variation, with no evolutionary fitness consequences or effects, in the lockstep runs.  Additionally, population statistics are identical and sensory input statistics are comparable between paired runs.

\subsection{Simulations and Data Acquisition}

A set of 10 paired simulations, differing only in initial random number seed, were run in natural-selection and lockstep modes; i.e., 20 simulations in all.  Each was run for 30,000 time steps.  As Polyworld is continuous rather than generational, determining the number of generations is non-trivial.  In the past estimates have been known only to fall within a large range.  A low estimate based on average lifespan (about 300 time steps) would be 100 generations.  A high estimate based on the minimum age of fecundity (25 time steps) would be 1,200 generations.  A newly implemented lineage tracer produces a more accurate estimate of about 400 generations.

The world is seeded with a uniform population of agents that have the minimum number of neural groups and nearly minimal neuron and synapse counts.  While predisposed to some potentially useful behaviors, such as running towards green (food) and away from red (aggressive agent behaviors; see \citep{Yaeger1994} for details on color use in Polyworld), these seed organisms are not a viable species.  That is, unless they evolve they cannot sustain their numbers through reproduction and will gradually die out. 

During simulations, the activation of every neuron in the brain of every agent is recorded at every time step.  These brain function recordings are grouped into arbitrary (here, 1,000) time step bins, for all agents that died during the specified interval.  Utility programs are then used to calculate the complexity, \textbf{C}, of the neural dynamics of every agent's complete lifespan (hence the requirement for the agent's death).  We then compute mean complexities for these binned populations of agents as a function of time.  Finally, we compute means and standard deviations of the population means for the multiple natural-selection and lockstep simulations as a function of time to study general evolutionary trends in complexity.

Complexity can be calculated across all neurons, just the input neurons, or just the ``processing" neurons (all neurons except inputs).  All complexities presented here are based on processing neurons.  (In general, there is little difference in complexity trends between all neurons and processing neurons.)  Complexity varies as evolution produces changes in the parts of the genome that specify the neural architecture.

\section{Results and Discussion}

\begin{figure*}[tb]
	\centering
		\includegraphics[width=0.8\textwidth,angle=0]{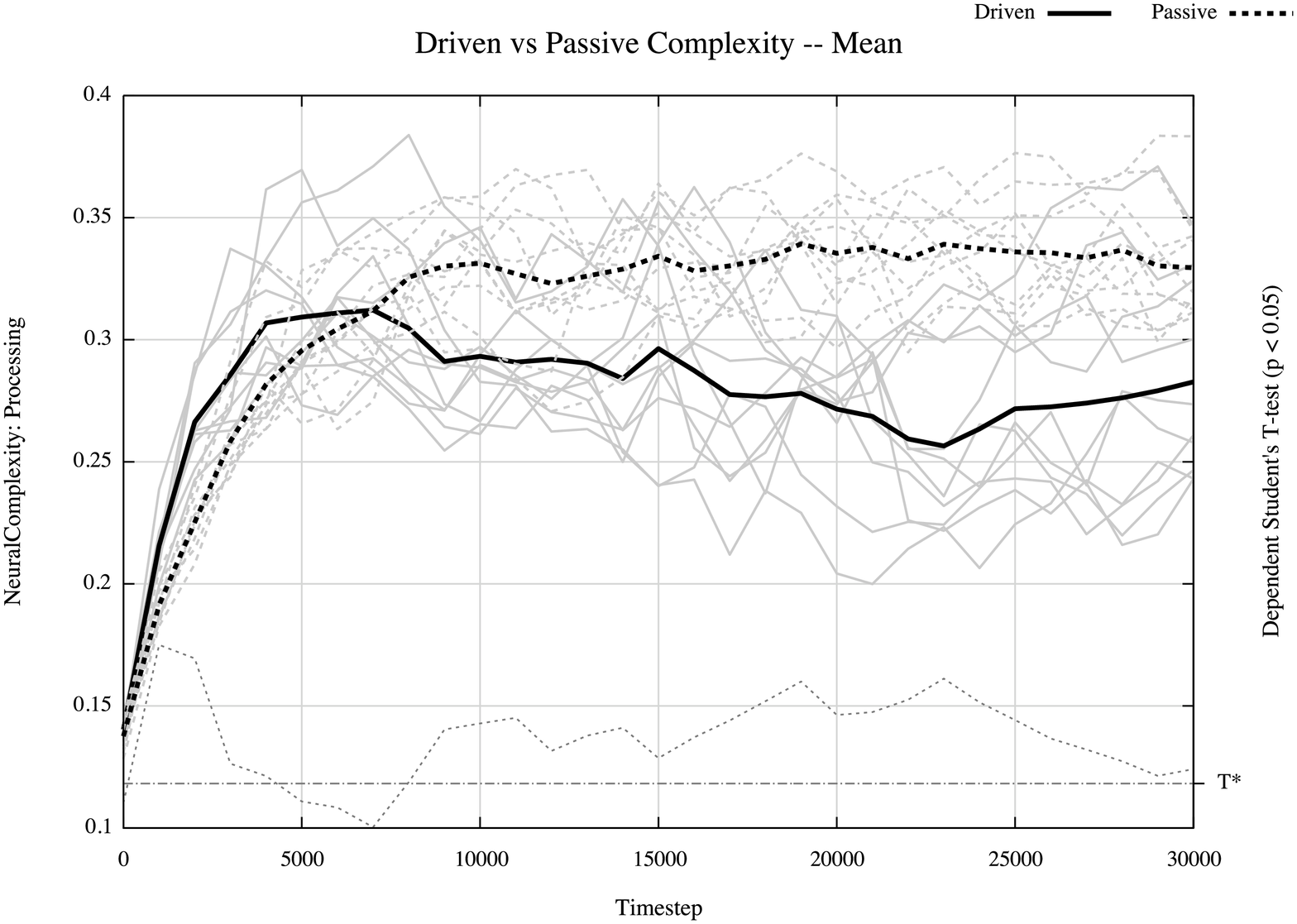}
		\caption{Driven and passive complexity \emph{vs.} time.  Light solid lines show population mean complexity for each driven, natural-selection run.  Light dashed lines show population mean complexity for each passive, lockstep run.  Heavy lines show means of all ten runs for corresponding line style.  Light dotted line at bottom shows dependent Student's T-test relative to horizontal T-critical line (labeled T*) for p~$>$~0.05.}
		\label{fig_cm}
\end{figure*}

Figure~\ref{fig_cm} shows complexity versus time for the previously described series of 10 paired driven (natural-selection) and passive (lockstep) simulations.  The lighter lines depict population means from individual runs.  The heavier lines depict means of all runs of a particular type (driven or passive).  Data is presented in this individual-plus-mean fashion, rather than mean-plus-standard-error fashion, to give a better feel for the nature of the variance between runs, and to identify some interesting events in a small number of the runs (discussed later).  Plotted beneath the complexity lines is a single dotted line that measures a paired or dependent Student's T-test computed on the same time interval as the complexity data are computed and plotted.  (A dependent test is used because these are paired runs with common initial conditions and enforced common population and genetic statistics.)  Where this line is above the horizontal T-critical (T*) line, this standard measure of statistical significance rejects the null hypothesis with p~$<$~0.05; at or below the T-critical line the null hypothesis cannot be rejected (at least not as reliably).  Given 10 pairs of runs, the number of degrees of freedom is 9, and T-critical is 1.833.

The first thing to note is a statistically significant faster growth rate in complexity in the driven runs than in the passive runs during approximately the first 4,000 time steps.  Evolution is clearly selecting for an increase in complexity during this early time period.  This makes sense intuitively since the seed population is known to be non-viable and must evolve or die out.  Increases in complexity during this period are of a distinct evolutionary advantage, producing descendant populations that are more capable of thriving in this particular environment than their ancestors.  During this time period, the evolution of complexity is clearly driven, with a bias towards increasing complexity.

\begin{figure*}[tbh!]
	\centering
		\includegraphics[width=0.8\textwidth,angle=0]{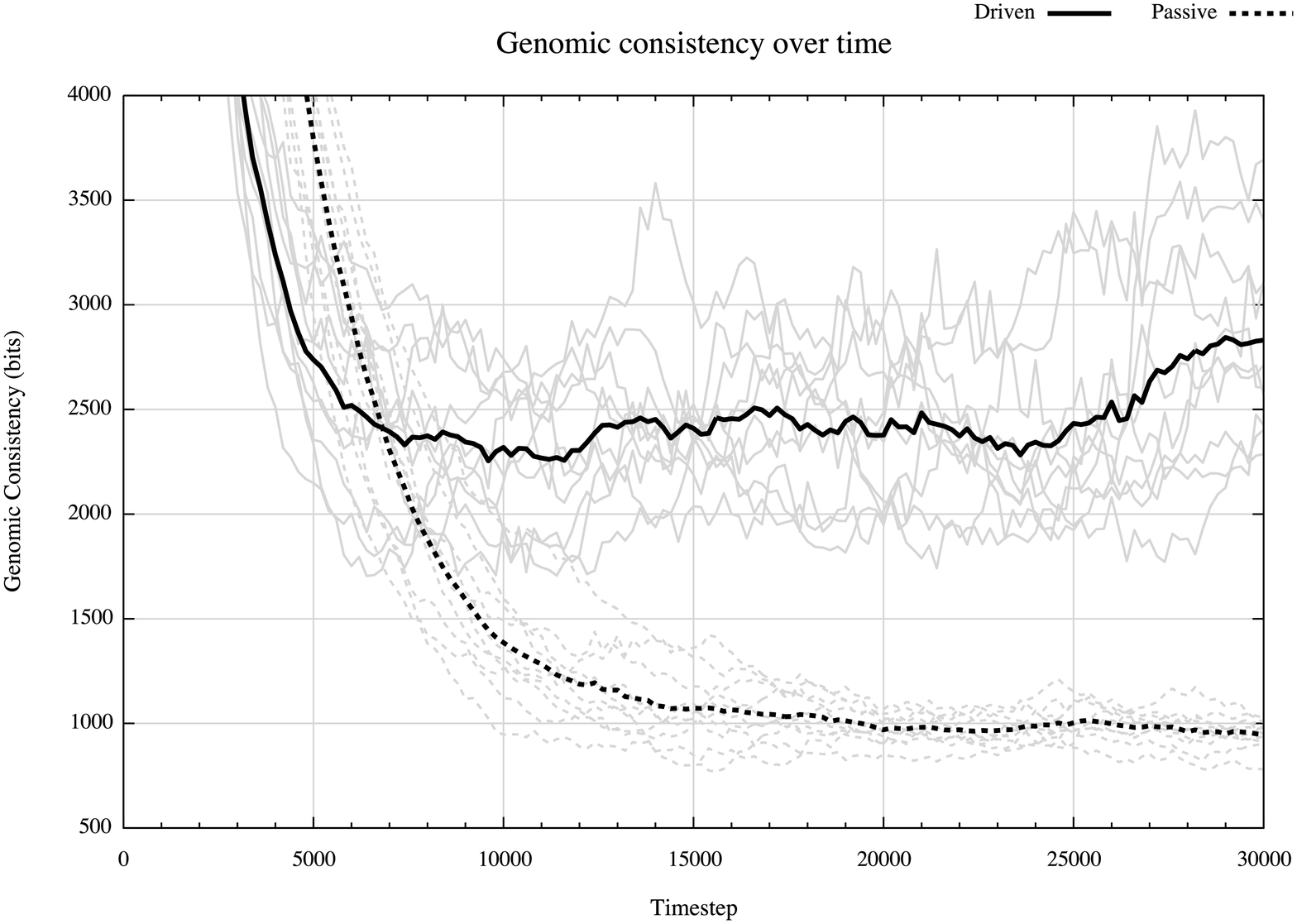}
		\caption{Genomic consistency \emph{vs.} time.  Light solid lines are for each driven, natural-selection run.  Light dashed lines are for each passive, lockstep run.  Heavy lines show means of all ten runs for corresponding line style.}
		\label{fig_gc}
\end{figure*}

The next thing to note is the early plateauing of complexity in the driven runs, allowing the randomly drifting complexity of the passive runs to catch and surpass them by around t=7,000.  This is the result of evolution having found a solution that is ``good enough", and the concomitant spread of the genes producing this solution throughout the population.  Seven out of the 10 natural selection simulations remain relatively stable around this modestly complex solution once it is found.  The intuition here is that any change away from this ``good enough" solution is likely to be detrimental, hence evolution selects for stability.  Note that this actively suppresses genetic drift and, indeed, a statistically significant difference between driven and passive runs, with passive complexity now being the larger of the two, is maintained from about t=8,000 to the end of the runs at t=30,000.

There is also a consistent, but less interesting, plateauing of complexity in the passive runs.  This is due solely to the individual bits of the underlying genome approaching a state of approximately 50\% on, 50\% off.  Effectively, the random walk has maximized variance as much as it can given the model parameters.  Though generally higher than the driven mean, complexities in the passive/random model are nowhere near the maximum obtainable with the full range of gene values (as observed in the complexity-as-fitness-function experiments discussed below); they just correspond to the range of complexities representable by the genome with an even mix of on and off bits.  Such larger values of complexity are potentially meaningful, but do not confer any evolutionary advantage on agents in these lockstep runs.

Finally, if one looks carefully at driven complexity for the individual runs, three (of the 10) runs make secondary transitions upward in complexity between t=20,000 and the end of the run, coincidentally reaching about the same level of complexity as the passive runs.  In this subset of runs, apparently a new or improved behavior emerges late in the simulation, and the genes producing this behavior spread throughout the population fairly rapidly.  Despite the simplicity of the world design used for these experiments, multiple viable, competing solutions have emerged and it is always possible that more solutions would emerge given time.  Importantly, for the future, it appears that this complexity measure provides a useful tool for ascertaining the onset of new, improved strategies, including speciation events, as well as a quantitative tool for assessing the neural changes that produced the new strategies.

\begin{figure}[tbh!]
  \centering
    \subfigure[]{\label{fig_hist-a}\includegraphics[width=0.4\textwidth]{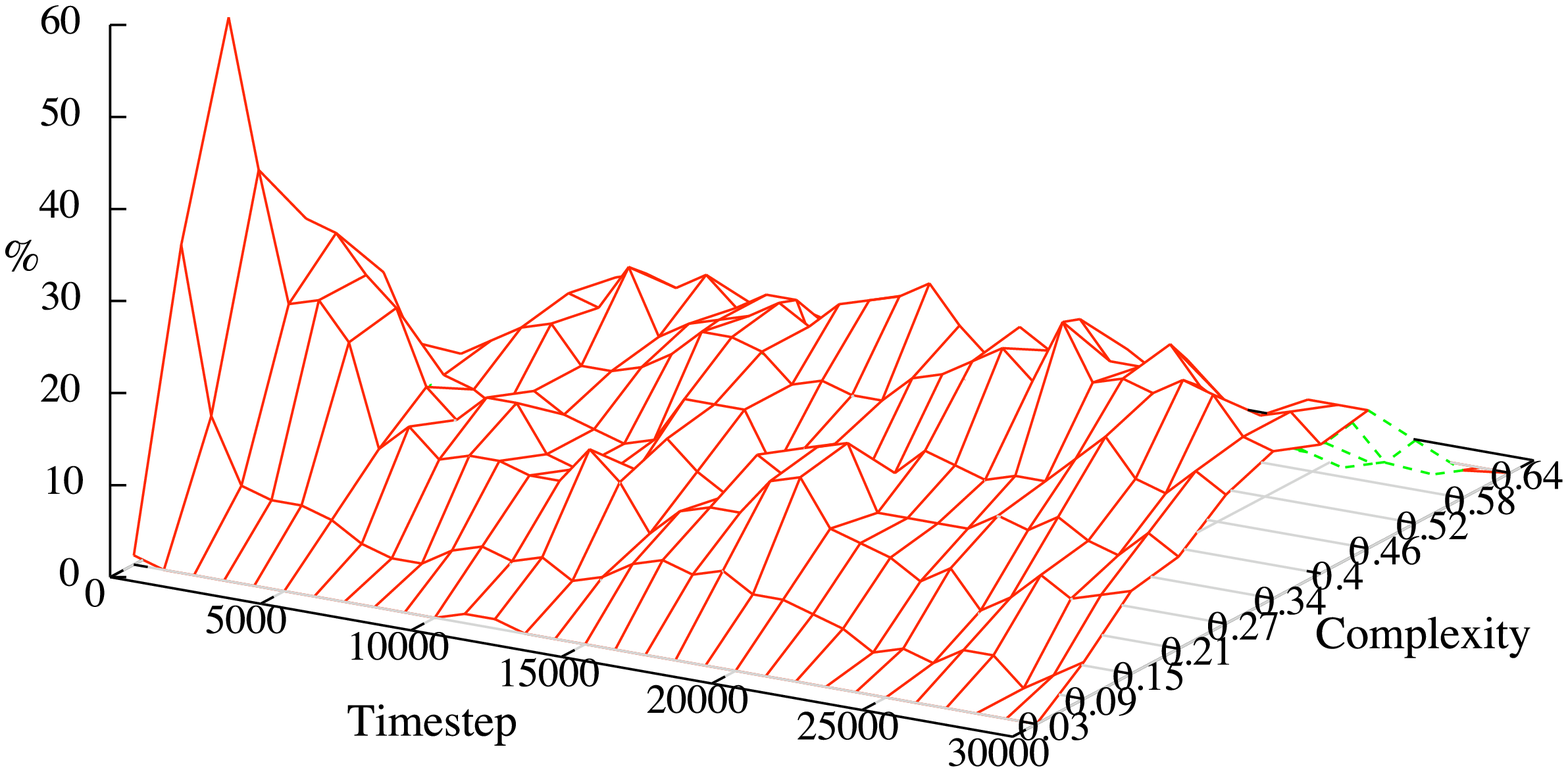}}
	\hspace{1cm}
	\subfigure[]{\label{fig_hist-b}\includegraphics[width=0.4\textwidth]{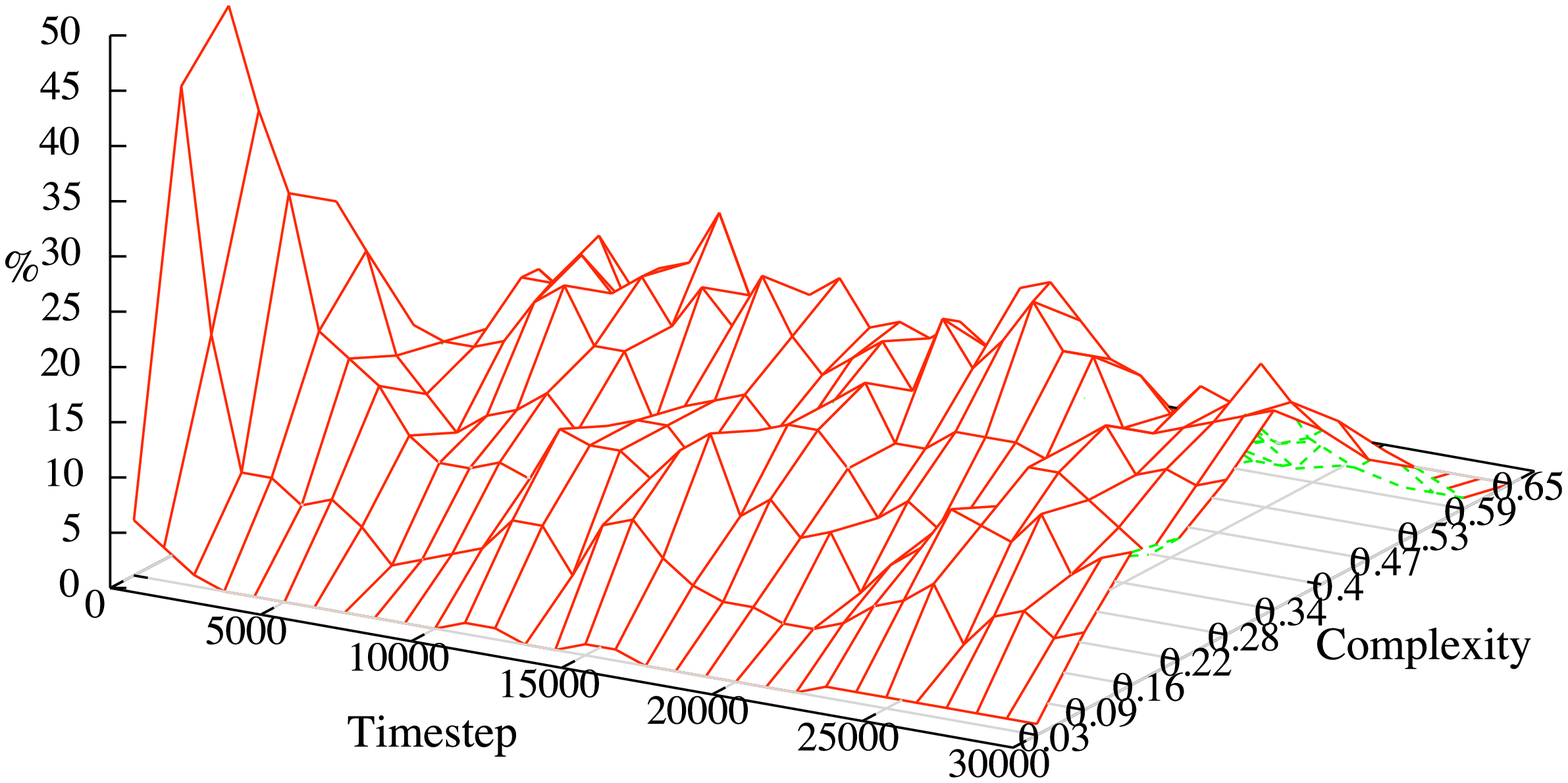}}
    \subfigure[]{\label{fig_hist-c}\includegraphics[width=0.4\textwidth]{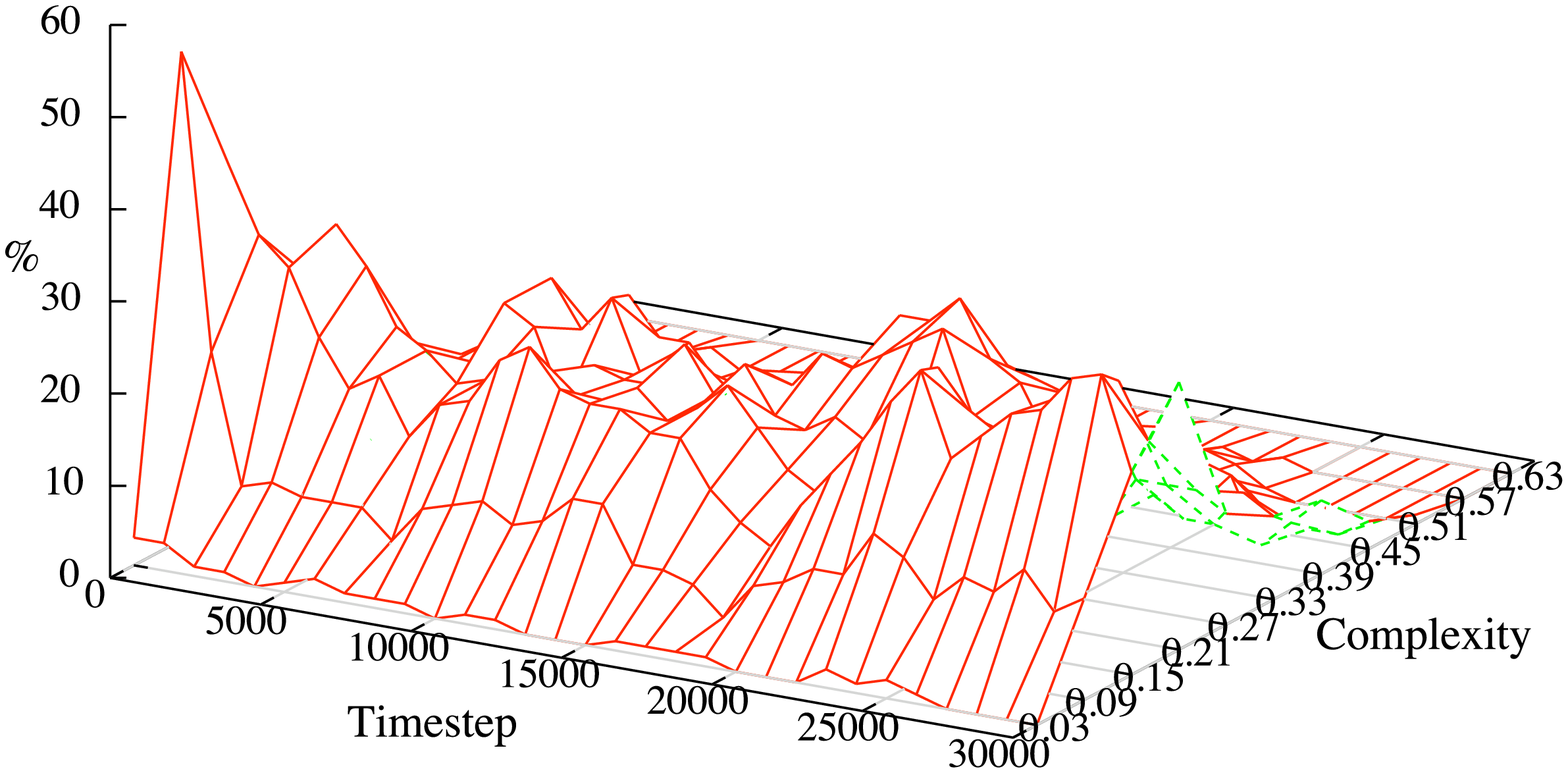}}
	\hspace{1cm}
    \subfigure[]{\label{fig_hist-d}\includegraphics[width=0.4\textwidth]{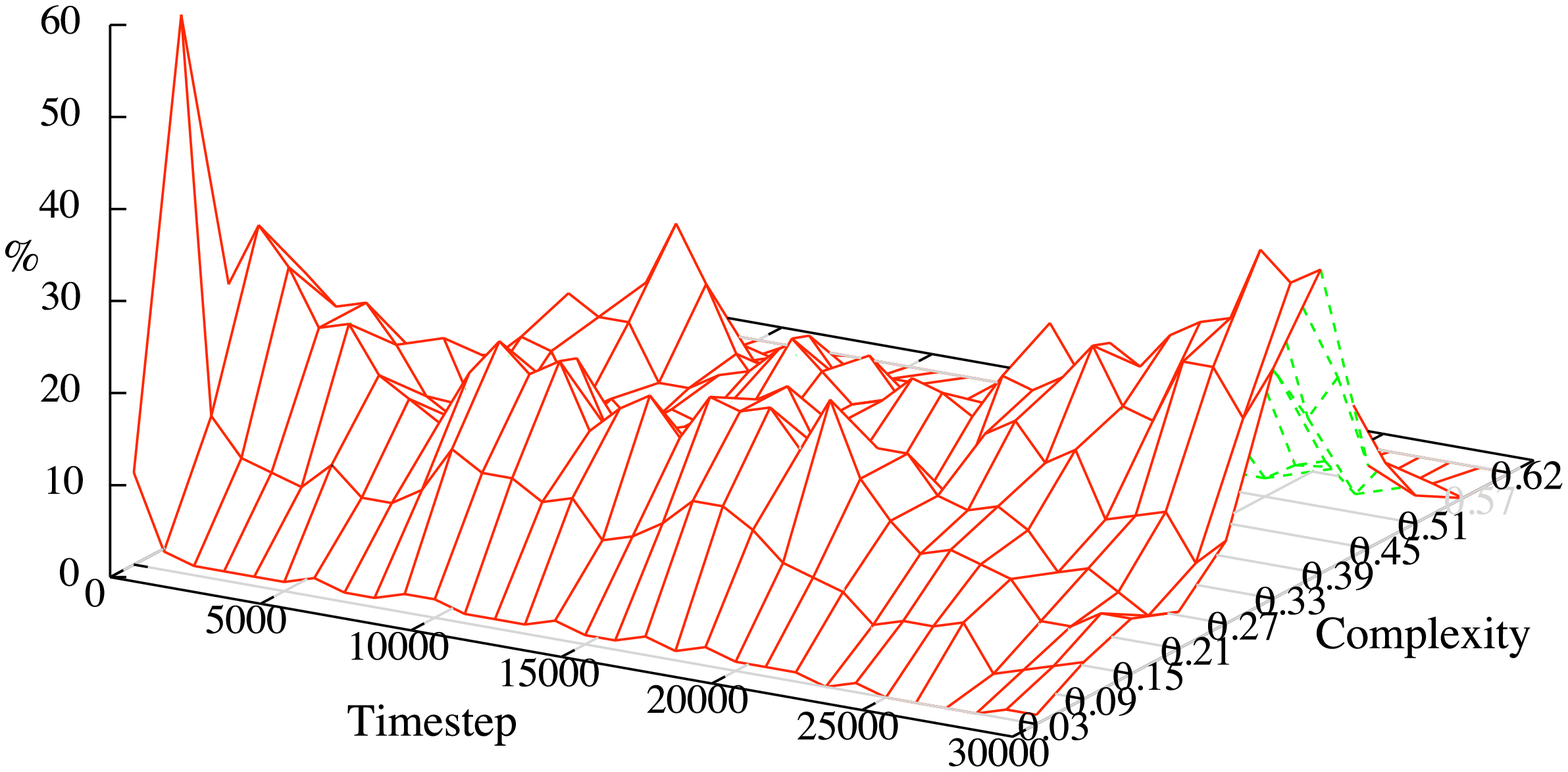}}
  \caption{Histograms of complexity over time for individual runs.  (a) and (b) are passive runs.  (c) and (d) are driven runs.}
  \label{fig_hist}
\end{figure}

Two of us [LY,OS] independently realized some years ago that if we were in a position to measure neural complexity in an artificial system, it might make sense to accelerate the evolution of complexity by using it as an explicit fitness function.  Although not elaborated upon here, preliminary experiments to this end have been carried out.  With the same values for the parameters controlling neural architecture, using complexity as a fitness function pushes mean complexity up to around 0.9, roughly three times the levels obtained by this series of driven (0.275) and passive (0.33) runs.  However, selecting purely for complexity consistently produces stereotypic spinning behaviors by the agents, that would not be of much value under natural selection, and we suspect are the result of a maximization of entropy and mutual information in the sensory inputs.  This differs from the direct coupling between complexity and behavior found in \citep{Sporns2006}, probably due to differences in the range of possible behaviors and the nature of the sensory inputs in the simulation environments used for these two studies.

\citet{Adami2000,Adami2002} has defined a measure of genomic complexity in terms of how much information a species' genome encodes about its environment, based on the cross-population entropy at each genomic site (here, individual bits).  The measure has acknowledged constraints---it only applies to a single species in a single, static niche, thus failing to capture issues related to biodiversity, environmental variability, or broader ecologies.  More generally, we suspect it may be a better measure of genomic consistency or specialization of a species than of complexity, but there is no question that the aggregate stability of gene sites in a species' population is an important measure of the success of that species at encoding information about its environment in its genome.  Since our current series of simulations are deliberately simple and may probably best be thought of as the evolution of a single (or at least highly related) species in a single, static niche, we decided to investigate the evolution of this genomic consistency.

Figure~\ref{fig_gc} shows the evolution of Genomic Consistency (GC) over time.  Since the world is seeded with a uniform population, GC is initially extremely large, as the measure effectively assumes the current genetic structure is an evolved response to the environment and perfect uniformity is maximally consistent.  (However, we do not feel this should be seen as maximally complex, hence our renaming of Adami's genomic complexity.)  Accordingly, the vertical extent of the graph has been truncated in order to focus on the more interesting results.  The main observation is the dramatic difference between driven and passive runs.  The passive runs produce extensive random gene edits, thus minimizing GC.  The driven runs demonstrate a larger, stabilized GC across the population over time due to natural selection for specific traits that increase the evolutionary fitness of the agents.  There is a hint of a modest upward trend in GC after it bottoms out around t=12,000, suggesting a possible continued incorporation of information about the environment in the genome of these agents, but so far no attempt has been made to establish statistical significance.

Another way to draw some understanding from this series of driven \emph{vs.} passive runs is to look at the distribution of complexity in the populations over time.  Figure \ref{fig_hist} shows time-series histograms of complexity throughout the population for four sample simulations---two passive, two driven.  The two passive runs show a generalized increase in variance, due to the diffusive random walk away from the low complexity of the seed agents.  The two driven runs show a more peaked distribution around the complexity values attained by the viable populations emerging as a result of natural selection.  Figure \ref{fig_hist-c} is representative of the majority of the driven runs, showing a shift towards a modest level of complexity.  Figure \ref{fig_hist-d} is representative of the small number of driven runs in which a secondary transition to a different behavior and higher level of complexity emerged late in the simulation.

\section{Conclusions}

We have demonstrated a technique for directly comparing and assessing neural complexity growth in equivalent driven and passive systems.  Using this technique we have shown evolutionary selection for \emph{increased} complexity, in a ``driven" fashion, as well as selection for complexity \emph{stability}.  Though we have not demonstrated it here, there is little doubt that a system in which the cost of neural complexity exceeded its value would result in a driven \emph{reduction} in complexity, the way dark dwelling organisms in a cave have been known to give up their eyes.  This paints a complex picture of evolutionary selection for increasing, stable, and decreasing complexity, none of which corresponds to a purely ``passive" mechanism of complexity change.  At this scale, evolution is entirely driven, with changes in complexity always being selected for or against.  Scale, however, is very important to this discussion.

\citet{Gould1996} and \citet{Dawkins1997} have argued strongly for passive and driven evolutionary trends, respectively.  However, much of the disconnect between them seems to be precisely an issue of scale.  Dawkins is unquestionably correct about evolution being driven on a short time scale, for a small set of related species.  Yet Gould \emph{may} be correct, as well, about evolution being fundamentally passive on a longer time scale, over the entire tree of biological life.  In one of the earliest works to model evolution computationally in order to characterize active versus passive trends, \citet{Raup1973} called attention to the fact that fully deterministic, driven trends acting at small scales are in fact likely to be at the base of larger scale trends, even if those large scale trends turn out to be passive.  A mix of many, potentially opposing trends might very well appear random and undirected when integrated together.  What our current simulations show is that, indeed, while evolution undoubtedly drives complexity changes, according to perfectly standard expectations about the evolutionary fitness of those changes, it does not drive in just one direction.  When complexity increase is of an evolutionary advantage it will be selected for, just as will complexity decrease.  And when a species' complexity is ``good enough", so that any increase or decrease is likely to involve a step away from a local fitness maximum, evolution will mildly select for and stabilize the existing level of complexity.  This goes a long way towards explaining the observation by \citet{Dennett1996}, ``The cheapest, least intensively designed system will be `discovered' first by Mother Nature, and myopically selected."

Looking forward, though we have yet to address the issue experimentally, we expect any increase in agent interactions with the world, any increase in complexity of the environment, and any increase in the available range of niches---\citet{Knoll2000}'s expanding \emph{ecospace}---to produce an increase in evolved neural complexity of agents in the world.  All niches are not created equal, and we suspect that evolutionary occupation of more and richer parts of ecospace will, as Knoll suggests, result in a fundamentally driven growth in complexity both at the largest scales of biology and in our artificial worlds.  And in Polyworld we expect \textbf{C}, our measure of neural complexity, to quantify and document that trend.

\section{Acknowledgements}

Thanks to David Brinda for Python and gnuplot hacking.

\footnotesize
\bibliographystyle{apalike}
\bibliography{trends}

\end{document}